\pdfoutput=1

\documentclass[11pt]{article}

\usepackage{acl}

\usepackage{times}
\usepackage{latexsym}

\usepackage[T1]{fontenc}

\usepackage[utf8]{inputenc}

\usepackage{microtype}

\usepackage{inconsolata}

\usepackage{booktabs}
\usepackage{graphicx}
\usepackage{amsmath}
\usepackage{multirow}
\usepackage{multicol}
\usepackage{tabularx}
\usepackage{caption}
\usepackage{xcolor}
\usepackage{colortbl}
\usepackage{makecell}
\usepackage{algorithm}
\usepackage{algorithmic}

\definecolor{myred}{rgb}{0.8, 0.25, 0.33}
\definecolor{myblue}{rgb}{0.33, 0.5, 0.8}
\definecolor{mygrey}{rgb}{0.94, 0.94, 0.94}

\newcolumntype{g}{>{\columncolor{mygrey}}c}

\title{Counterfactual Debiasing for Generating Factually Consistent \\ Text Summaries}

\author{Chenhe Dong$^{1,}$\footnotemark[1] \quad Yuexiang Xie$^{2}$ \quad Yaliang Li$^{2}$ \quad Ying Shen$^{1}$\\
$^{1}$Sun Yat-sen University \quad $^{2}$Alibaba Group \\ 
dongchh@mail2.sysu.edu.cn \quad
\{yuexiang.xyx, yaliang.li\}@alibaba-inc.com \\
sheny76@mail.sysu.edu.cn \\
}

\begin{document}
\maketitle
\renewcommand*{\thefootnote}{\fnsymbol{footnote}}
\footnotetext[1]{Work done at Alibaba.}
\renewcommand*{\thefootnote}{\arabic{footnote}}

\begin{abstract}
Despite substantial progress in abstractive text summarization to generate fluent and informative texts, the factual inconsistency in the generated summaries remains an important yet challenging problem to be solved. In this paper, we construct causal graphs for abstractive text summarization and identify the intrinsic causes of the factual inconsistency, i.e., the language bias and irrelevancy bias, and further propose a debiasing framework, named \textsc{CoFactSum}, to alleviate the causal effects of these biases by counterfactual estimation. Specifically, the proposed \textsc{CoFactSum} provides two counterfactual estimation strategies, i.e., Explicit Counterfactual Masking with an explicit dynamic masking strategy, and Implicit Counterfactual Training with an implicit discriminative cross-attention mechanism. Meanwhile, we design a Debiasing Degree Adjustment mechanism to dynamically adapt the debiasing degree at each decoding step. Extensive experiments on two widely-used summarization datasets demonstrate the effectiveness of \textsc{CoFactSum} in enhancing the factual consistency of generated summaries compared with several baselines.
\end{abstract}

\section{Introduction}
Abstractive text summarization~\cite{dong2019unified,lewis2020bart,zhang2020pegasus} has witnessed great success in generating remarkably fluent and diversified summaries that approach human-level performance, which is inextricably tied to the usage of large pertained language models.
Nevertheless, the generated summaries often contain factually inconsistent errors against the source documents~\cite{narayan2018dont-give,maynez2020on-faithfulness}. 
For example, as shown in Figure~\ref{fig:example}, the subject is incorrectly predicted as ``a 10-year-old boy'' instead of the correct subject ``a cricket team'', and the team's final score is wrongly predicted as ``one'' instead of ``zero''. 
Such inconsistencies in the generated summaries can mislead and confuse the public and even arise legal and moral risks,  which brings significant rectification costs and limits the applications of abstractive text summarization.

To tackle such factually inconsistent issues, several techniques have been proposed in recent years, which can be divided into three categories: 
(i) \textit{fact encoding}, which integrates additional fact-related information during encoding~\cite{huang2020knowledge,zhu2021enhancing}; 
(ii) \textit{post editing}, which adopts a rectification model to correct the generated summaries~\cite{dong2020multi-fact,cao2020factual}; 
and (iii) \textit{auxiliary loss applying}, which designs an auxiliary loss to penalize the model for generating factually inconsistent texts~\cite{li2020dont,cao2021cliff}. 
However, most of these existing studies neglect the intrinsic causes of the factual inconsistency in abstractive text summarization.

\begin{table}
\centering
\resizebox{\columnwidth}{!}{
\begin{tabularx}{1.25\columnwidth}{p{1.2\columnwidth}}
\toprule
\textbf{Source document}: \textcolor{myblue}{No batsman from Bapchild Cricket Club was able to get off the mark} against Christ Church University in Canterbury. ``We couldn't believe it, \textcolor{myred}{all they needed to do was hit a wall to get one run},'' Christ Church player Mike Rose told the Crawley Observer. \textcolor{myblue}{Somerset club Langport set the record for the lowest score when they were dismissed for zero in 1913.} Wirral CC were bowled out for three in a Cheshire League Division Three fixture in 2014... \\

\midrule
\textbf{Factually consistent summary}: \textcolor{myblue}{A cricket team was bowled out for 0} in just 20 balls in a county six-a-side indoor championships match. \\

\midrule
\textbf{Factually inconsistent summary}: \textcolor{myred}{A 10-year-old boy} has broken the record for the lowest score ever made in first-class cricket when \textcolor{myred}{he hit one run} in his first match. \\

\bottomrule
\end{tabularx}}
\vspace{-0.1in}
\captionof{figure}{An example of a factually inconsistent summary. The \textcolor{myblue}{supporting and consistent facts} in the source document and summary are marked in \textcolor{myblue}{blue}, and \textcolor{myred}{the irrelevant and inconsistent facts} are marked in \textcolor{myred}{red}.}\label{fig:example}
\vspace{-0.1in}
\end{table}

Considering the generation process of abstractive text summarization models, the generated summaries rely on two key factors: the language prior knowledge acquired during pre-training, and the information contained in the source document, both of which contribute to the fluency and informativeness of generated summaries. 
However, they might also introduce \textit{language bias} and \textit{irrelevancy bias} which result in factual inconsistency.

To be more specific, the language prior knowledge is learned from a large number of corpora during the pretraining process, which can be a double-edged sword for text summarization. On the one hand, it helps text summarization models to generate fluent texts; On the other hand, it causes hallucinations due to the inevitably introduced spurious linguistic correlations, as discussed in previous studies~\cite{niu2021counterfactual-vqa,qian2021counterfactual,xie2021factual}.
Besides, due to the long source documents, irrelevant information (e.g., mismatched and uncorrelated entities) also plays a non-negligible role in causing factual inconsistency. 
For example, as shown in Figure~\ref{fig:example}, the unfaithful content ``he hit one run'' is likely to be inferred from mismatched tokens ``hit a wall to get one run'' in the source document, which is also about the topic of team score but does not correspond to the ground-truth subject.

Shed light on the above insights,  we incorporate the idea of causal inference~\cite{pearl2001direct,pearl2018the-book} into text summarization to ensure the factual consistency of the generated summaries by eliminating the language and irrelevancy biases.
Firstly, we build up a causal graph among the language prior knowledge, the important and irrelevant information of the source document, and the generated summary to demonstrate their causal relationships in abstractive text summarization.
Then, based on the causal graph, we propose a \textbf{Co}unter\textbf{Fact}ual debiasing framework for abstractive \textbf{Sum}marization, named \textbf{\textsc{CoFactSum}}, to estimate and alleviate the causal effects of language and irrelevancy biases on the generated summary.
The proposed \textsc{CoFactSum} consists of two counterfactual estimation strategies, including Explicit Counterfactual Masking (ECM) with an \textit{explicit} dynamic masking strategy, and Implicit Counterfactual Training (ICT) with an \textit{implicit} discriminative cross-attention mechanism.
Furthermore, we design a Debiasing Degree Adjustment (DDA) module to dynamically adapt the debiasing degree at each decoding step, improving the ability of the proposed framework to position the factual inconsistencies in the generated summaries.

We conduct a series of experiments on two widely adopted abstractive summarization datasets, i.e., CNN/DM~\cite{hermann2015teaching} and XSum~\cite{narayan2018dont-give} to show the effectiveness of \textsc{CoFactSum}.
The experimental results demonstrate that \textsc{CoFactSum} achieves significant improvement in generating factually consistent summaries compared to baseline models.
\section{Methodology}
In this section, we first introduce how we build up the causal graph for abstractive text summarization in \S\ref{ssec:causal_graph}, and conduct causal effect estimation in \S\ref{ssec:effect_estimation}. Then we describe the instantiation strategies of \textsc{CoFactSum} in \S\ref{ssec:instantiation}.

\subsection{Causal Graph Construction}\label{ssec:causal_graph}
The causal graph of abstractive text summarization can be given as a directed acyclic graph $\mathcal{G} = \{\mathcal{V}, \mathcal{E}\}$, which represents the causal relationships (i.e., $\mathcal{E}$) between different variables (i.e., $\mathcal{V}$). 
The causal graph consists of five variables: the source document $X$, the important information $U$, the irrelevant information $R$, the language prior $P$, and the generated summary $Y$, as shown in Figure~\ref{fig:causal_graph} (a). 
The observed values of these variables are denoted by the corresponding lowercase letters.

Since the source document $X$ is composed of the important information $U$ and the irrelevant information $R$, their causal relationships are denoted by the paths $X \to U$ and $X \to R$, respectively. 
During the generation process, the generated summary $Y$ is directly affected by the important information $U$, the irrelevant information $R$, and the language prior $P$, which can be expressed by $U \to Y$, $R \to Y$, and $P \to Y$. 
Note that the effect of language prior knowledge $P$ and irrelevant information $R$ on the generated summary $Y$ (i.e., $R \to Y$ and $P \to Y$) might introduce language bias and irrelevancy bias, which is considered to be estimated and mitigated in this study.

\subsection{Causal Effect Estimation}\label{ssec:effect_estimation}
Based on the causal graph, we can estimate the causal effects of language bias and irrelevancy bias on the generated summary.

\begin{figure}
\includegraphics[trim=10 300 300 10, width=0.92\columnwidth, clip]{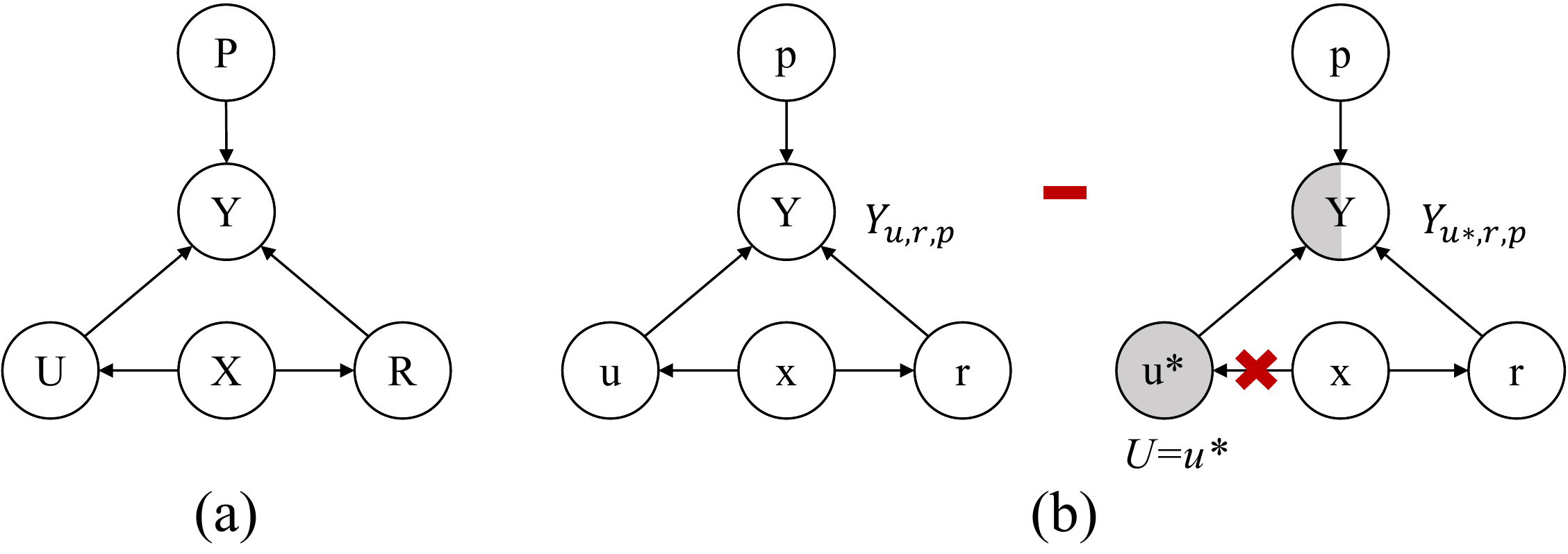}
\vspace{-0.1in}
\centering
\caption{Illustration for (a) the basic causal graph and (b) our debiasing framework \textsc{CoFactSum}.}\label{fig:causal_graph}
\vspace{-0.05in}
\end{figure}

\paragraph{Total Effect}
In the causal graph, suppose that the document $X$ is set to $x$, the underlying important and irrelevant information $U$ and $R$ is set to $u$ and $r$, respectively, and the language prior $P$ is set to $p$, then the generated summary $Y$ can be given as:
\begin{equation}
\begin{aligned}
Y_{u, r, p} &= Y(do(U=u), do(R=r), do(P=p)) \\
&= Y(U=u, R=r, P=p),
\end{aligned}
\end{equation}
where the $do$ operator can be omitted according to the back-door criteria~\cite{pearl2009causal}. 
To measure the total effect on $Y$, we need to compare the potential outcomes of the same individual under the treatment and no-treatment conditions, where the no-treatment condition can be approximated by setting $U, R, P$ to empty values $u^{*}, r^{*}, p^{*}$ under the counterfactual scenario. 
Formally, the total effect can be given as:
\begin{equation}
E_{total} = Y_{u, r, p} - Y_{u^{*}, r^{*}, p^{*}}.
\end{equation}

\paragraph{Bias Elimination}
Similarly, the causal effects of language prior knowledge $P$ and irrelevant information $R$ on the generated summary $Y$ can be estimated as:
\begin{equation}
E_{bias} = Y_{u^{*}, r, p} - Y_{u^{*}, r^{*}, p^{*}},
\end{equation}
where we set $U=u^{*}$ to exclude the causal effect of the important information $U$ on $Y$.
To eliminate the language bias and irrelevancy bias in the generation process, we remove their causal effects on the generated summary from the total effect. Formally, it can be given as:
\begin{equation}
E_{total} - E_{bias} = Y_{u, r, p} - Y_{u^{*}, r, p}.\label{eq:remove_bias}
\end{equation}
The equation can also be regarded as the estimation of the causal effect of important information $U$ on the generated summary $Y$ when given the $R=r$ and $P=p$, as illustrated in Figure~\ref{fig:causal_graph} (b).

\subsection{Instantiation}\label{ssec:instantiation}
In order to instantiate Equation (\ref{eq:remove_bias}) in abstractive text summarization, the first term, i.e., $Y_{u, r, p}$, can be interpreted as the generated summaries of conventional abstractive text summarization models. For the second term, i.e., $Y_{u^{*}, r, p}$, we design two counterfactual strategies, i.e., Explicit Counterfactual Masking (ECM) and Implicit Counterfactual Training (ICT), which are designed for estimating during the inference process and optimizing during the training process, respectively.

\paragraph{Explicit Counterfactual Masking (ECM)}
Previous studies~\cite{xie2021factual} have used masking techniques to block the causal effect of important information on the generated summary. 
However, the proposed ECM is different from previous studies in that it considers that during the generation process, the decoder attends to different tokens of the source document at different decoding steps. 
For example, when generating the team score in Figure~\ref{fig:example}, the tokens related to numbers in the source document are more likely to catch the attention of the decoder.
Therefore, we propose to dynamically determine the important tokens in the source document w.r.t. each generated token, rather than using a fixed set of important tokens.

Specifically, we use the cross-attention score as an indicator and employ a top-$K$ strategy to pick up the top $K$ positions with the maximum scores as the important positions. 
To remove these important tokens from the source document without causing the disparity between training and inference, we use a special token ``[MASK]'' to explicitly replace the important tokens, similar to the pre-training stage of most Transformer-based language models~\cite{devlin2019bert,zhang2020pegasus}.
We also adopt a debiasing ratio $\alpha$ ($\alpha \leq 1$) to adjust the extent of debiasing, in order to preserve the informativeness of generated summaries. 
Formally, the probability of each generated token $y_t$ with ECM can be given as:
\begin{equation}
\Pr(y_t|x) = \Pr(y_t|y_{<t}, x;\theta) - \alpha\cdot \Pr(y_t|y_{<t}, x';\theta),
\end{equation}
where $x'$ denotes the masked document, and $\theta$ denotes the model parameters.

\paragraph{Implicit Counterfactual Training (ICT)}
In addition to ECM, which is used during inference, a counterfactual training strategy with a discriminative cross-attention mechanism is further proposed to implicitly minimize the causal effect of bias on the generated summaries.

Specifically, at each decoding step, the source document is dynamically split into two disjoint partitions (i.e., important tokens $x_u$ and irrelevant tokens $x_r$) based on cross-attention scores. Then the decoder model separately attends to these partitions for counterfactual training. The probability of each generated token $y_t$ at decoding step $t$ can be represented as $\Pr(y_t | y_{<t}, x_u; \theta')$ and $\Pr(y_t | y_{<t}, x_r; \theta')$, respectively, where $\theta'$ denotes the parameters of the counterfactual summarization model.

Intending to guide the counterfactual text summarization model to rely less on the important tokens, we use an unlikelihood loss $\mathcal{L}_{unl}$ to penalize the sequence log-likelihood when the model attends to important tokens:
\begin{equation}
\mathcal{L}_{unl} = - \sum_{t = 1}^{|y|} \log \left(1 - \Pr(y_t | y_{<t}, x_u; \theta')\right),
\end{equation}
where $y$ is the ground truth summary.
Meanwhile, a cross-entropy loss $\mathcal{L}_{xent}$ is adopted to increase the probabilities of tokens that are generated when attending to irrelevant tokens:
\begin{equation}
\mathcal{L}_{xent} = - \sum_{t = 1}^{|y|} \log \Pr(y_t | y_{<t}, x_r; \theta').
\end{equation}

Moreover, we adopt a Kullback-Leibler (KL) divergence loss $\mathcal{L}_{kl}$ to further push away the predicted distributions over vocabulary when attending to the important tokens and irrelevant tokens respectively, which can be formally given as:

\begin{small}
\begin{equation}
\mathcal{L}_{kl} = - \sum_{t = 1}^{|y|} \operatorname{KL} \left(\Pr(\cdot|y_{<t}, x_u; \theta') || \Pr(\cdot | y_{<t}, x_r; \theta')\right).
\end{equation}
\end{small}
Finally, the training loss can be defined by:
\begin{equation}\label{eq:total_loss_cross_attn}
\mathcal{L} = \mathcal{L}_{unl} + \gamma \mathcal{L}_{xent} + \lambda \mathcal{L}_{kl},
\end{equation}
where $\gamma, \lambda$ are hyperparameters to control the strength of adopted loss functions. 
Note that only the parameters of the decoder are updated while the word embeddings and the encoder are frozen. The reason is that the debiasing process happens during the decoding stage and the encoder outputs in the treatment and no-treatment conditions should be the same. 

Applying the above counterfactual training process, we train a counterfactual decoder as an instantiation of $Y_{u^{*},r,p}$. 
The debiased probability of each generated token $y_t$ with ICT is given as:
\begin{equation}
\Pr(y_t|x) = \Pr(y_t|y_{<t}, x; \theta) - \beta\cdot \Pr(y_t|y_{<t}, x; \theta'),
\end{equation}
where $\beta$ ($\beta \leq 1$) is a hyperparameter.

\paragraph{Debiasing Degree Adjustment (DDA)}
Taking both ECM and ICT into consideration, we point out that debiasing at every decoding step to the same extent might not be an optimal solution, since the intermediately generated sentences at different decoding steps have different factually inconsistent degrees.
It is reasonable to conduct more intensive debiasing when the generated sentence is relatively less consistent and vice versa. 

To this end, we propose a dynamic adjustment strategy for the debiasing degrees at different decoding steps. We first pre-train a factual consistency predictor based on synthetic factually inconsistent summaries. Then the debiasing ratio at each decoding step can be dynamically adapted according to the predicted factually inconsistent score. In detail, the prediction process is formulated as a sequence labeling task. The label of each token in the factually inconsistent summaries is obtained by comparing each sample with its corresponding gold summary, where the mismatched tokens are labeled as \textit{inconsistent} and others as \textit{consistent}.

During training at the $t$-th decoding step, the predictor receives the following four representations: the original decoding hidden states $\mathbf{h}_t \in \mathcal{R}^d$, the counterfactual hidden states generated from the masked source document $\mathbf{h}^{\prime}_t \in \mathcal{R}^d$, and the element-wise multiplication and difference of the above two hidden states. These representations are concatenated and sent to a fully connected layer and a softmax function to obtain the predicted scores, as formulated by:
\begin{gather}
\mathbf{S}_t = \operatorname{softmax}(\mathbf{W} \cdot \mathbf{z}_t + \mathbf{b}) \in \mathcal{R}^2, \\
\mathbf{z}_t = [\mathbf{h}_t; \mathbf{h}^{\prime}_t; \mathbf{h}_t \odot \mathbf{h}^{\prime}_t; \mathbf{h}_t - \mathbf{h}^{\prime}_t] \in \mathcal{R}^{4d},
\end{gather}
where $d$ is the dimension of hidden states, $\mathbf{W} \in \mathcal{R}^{2 \times 4d}, \mathbf{b} \in \mathcal{R}^{2}$ are learnable parameters in the linear layer, $[;]$ denotes the concatenation, $\odot$ is the element-wise multiplication, and $\mathbf{S}_t$ contains the factually consistent score $S_t^{c}$ and factually inconsistent score $S_t^{ic}$ in which $S_t^{c} + S_t^{ic} = 1$. We use the cross-entropy loss to train the predictor and freeze all the parameters of the original text summarization model.

\begin{figure}
\includegraphics[trim=10 0 10 0, width=0.88\columnwidth, clip]{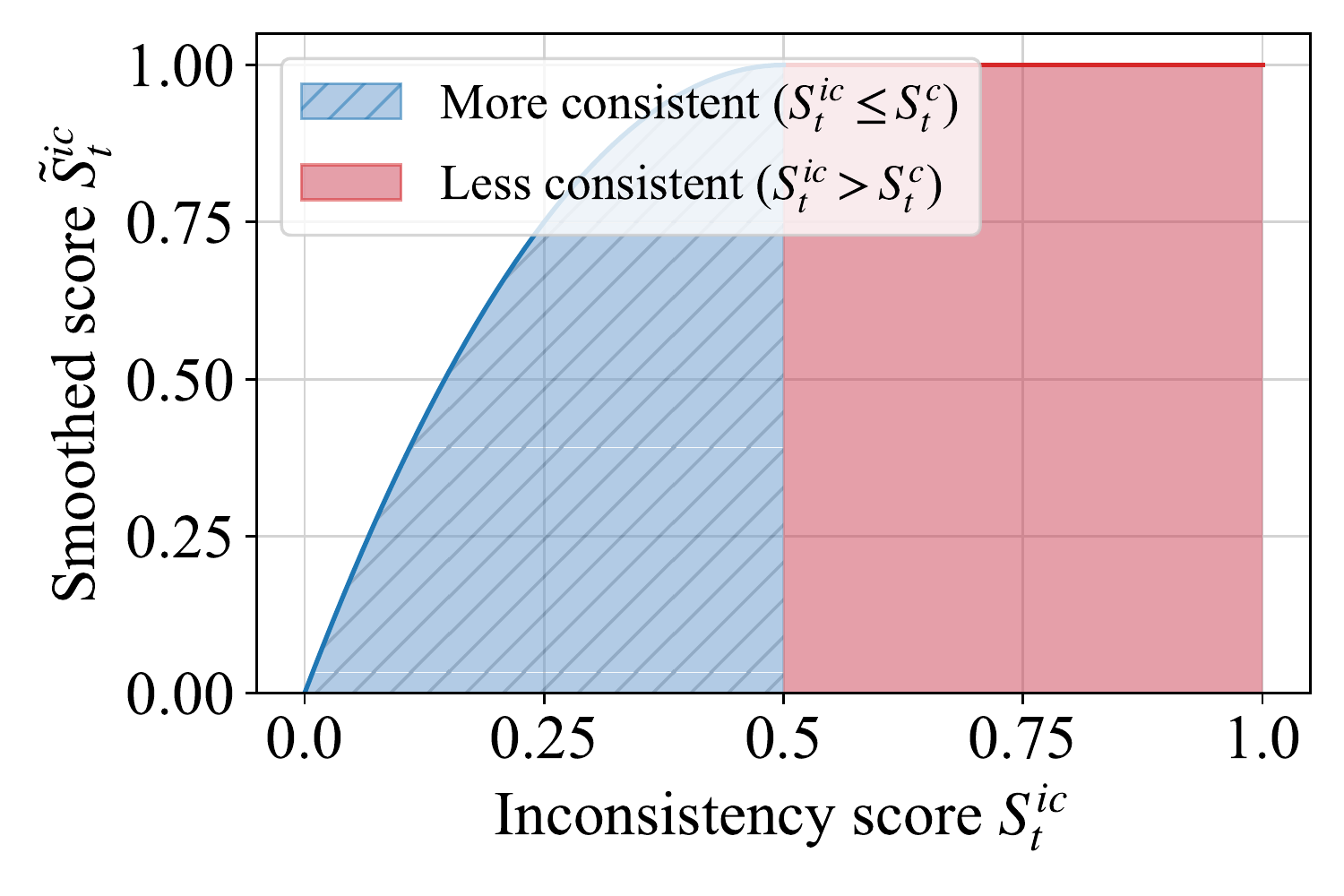}
\vspace{-0.2in}
\centering
\caption{The smoothing function used in DDA for the factually inconsistent scores.}\label{fig:smooth_function}
\vspace{-0.1in}
\end{figure}

\begin{figure}[h]
\vspace{-0.15in}
\begin{algorithm}[H]
\caption{\textsc{CoFactSum} Algorithm}
\label{algorithm1}
\begin{algorithmic}[1]
    \REQUIRE Source document $x$, original summarization model $f_\theta$, counterfactual summarization model $f^{\prime}_{\theta^{\prime}}$ trained with ICT, factual consistency predictor $g$ trained with DDA, maximum decoding step $T$
    \ENSURE Factually consistent summary $y$

    \STATE Initialize $y \gets \{\}$;
    \FOR {$t \gets 1$ to $T$}
        \STATE Feed $x, y_{<t}$ into $f$ to generate the probability of each token $y_t$ at $t$-th decoding step $\Pr(y_t | y_{<t}, x; \theta)$;
        \STATE Mask $x$ according to the cross-attention score to produce $x^{\prime}$;
        \STATE Feed $x^{\prime}, y_{<t}$ into $f$ to generate the probability $\Pr(y_t | y_{<t}, x^{\prime}; \theta)$;
        \STATE Feed $x, y_{<t}$ into $f^{\prime}$ to generate the probability $\Pr(y_t | y_{<t}, x; \theta^{\prime})$;
        \STATE Feed $x, x^{\prime}$ into $g$ to generate the smoothed factually inconsistent score $\tilde{S}^{ic}$;
        \STATE Calculate $\Pr(y_t | x)$ according to Equation (\ref{eq:final_eq}) and select $y_t^*$ with highest probability;
        \STATE $y \gets y \cup y_t^*$;
    \ENDFOR
    \RETURN $y$
\end{algorithmic}
\end{algorithm}
\vspace{-0.4in}
\end{figure}

During inference, we multiply the subtracted terms $\alpha\cdot \Pr(y_t|y_{<t}, x';\theta)$ and $\beta\cdot \Pr(y_t|y_{<t}, x; \theta')$ by a predicted factually inconsistent score to dynamically control the debiasing degrees. 
Besides, as we observed in our experiments, the factually inconsistent scores tend to vary dramatically across different decoding steps, thus we design a smoothing function to restrict their variation range and stabilize the inference. 
The overall predicted probability with debiasing can be formally given as:

\begin{small}
\begin{align}\label{eq:final_eq}
\Pr(y_t|x) = \Pr(y_t|y_{<t}, x;\theta) & - \tilde{S}^{ic}\cdot \Big(\alpha\cdot \Pr(y_t|y_{<t}, x';\theta)\nonumber\\
& + \beta\cdot \Pr(y_t|y_{<t}, x; \theta')\Big).
\end{align}
\end{small}

\noindent $\tilde{S}^{ic}$ is the smoothed factually inconsistent score, and at the $t$-th decoding step, it is calculated by:
\begin{equation}
\tilde{S}_t^{ic} = \left\{
\begin{array}{ll}
1 - (2S_t^{ic} - 1)^2, & S_t^{ic} \leq S_t^c \\
1, & S_t^{ic} > S_t^c
\end{array}
\right.,
\end{equation}
which is illustrated in Figure~\ref{fig:smooth_function}.
The overall training procedure of the proposed \textsc{CoFactSum} is summarized in Algorithm~\ref{algorithm1}.

\section{Experiment}
\begin{table*}
\caption{Automatic evaluation results on CNN/DM and XSum. Methods with $\dagger$ are conducted with released codes. \textbf{Bold} and \underline{underlined} indicate methods with the best and second-best performances, respectively. Columns in grey indicate metrics in terms of factual consistency.}\label{tab:main_results}
\centering
\vspace{-0.1in}
\setlength\tabcolsep{4pt}
\resizebox{\textwidth}{!}{
\begin{tabular}{lcgggggccgggggc}

\toprule
\multirow{2}{*}{\textbf{Methods}}& \multicolumn{7}{c}{\textbf{CNN/DM}}& \multicolumn{7}{c}{\textbf{XSum}} \\
\cmidrule(lr){2-8} \cmidrule{9-15}
~& \textbf{R-L}& \textbf{QAFE}& \textbf{QAGS}& \textbf{FCC}& \textbf{FT-C}& \textbf{FT-O}& \textbf{AVG}&   \textbf{R-L}& \textbf{QAFE}& \textbf{QAGS}& \textbf{FCC}& \textbf{FT-C}& \textbf{FT-O}& \textbf{AVG} \\
\midrule
\textsc{Pegasus}~\cite{zhang2020pegasus}&  \underline{40.48}& \underline{89.25}& 75.52& \underline{39.43}& 53.64& 67.86& \underline{52.81}&   \textbf{39.06}& 41.49& 21.47& \underline{25.29}& 6.17& \underline{3.72}& \underline{29.34} \\
\textsc{Unl}~\cite{li2020dont}&  39.15& 86.71& 74.72& 36.76& 53.31& 67.86& 51.51&   34.03& 38.51& 18.87& \textbf{25.92}& 4.45& 1.17& 25.91 \\
\textsc{Corr}~\cite{cao2020factual}&  39.79& 82.30& 69.49& 22.68& 49.46& 58.87& 48.18&   \underline{38.95}& 41.72& 21.73& 25.01& 6.10& 3.69& 29.30 \\
\textsc{CCGS}~\cite{chen2021improving}$\dagger$&  40.40& 87.24& 73.35& 37.09& \underline{54.71}& 67.40& 52.18&   38.68& 41.08& 21.14& 25.11& \underline{8.31}& 3.67& 29.27 \\
\textsc{CLIFF}~\cite{cao2021cliff}&  39.47& 88.64& \textbf{76.59}& 39.22& 54.57& \underline{71.02}& 52.74&   38.14& \textbf{43.34}& \underline{22.80}& 24.73& 6.24& 3.15& 29.10 \\
\textsc{SC}~\cite{xiao2022entity-based}$\dagger$& \textbf{41.34}& 82.45& 70.17& 30.15& 45.95& 52.12& 48.75&   38.34& 37.20& 19.87& 23.49& 4.76& 1.54& 27.86 \\
\midrule
\textsc{CoFactSum} (Ours)&  39.94& \textbf{90.18}& \underline{75.94}& \textbf{43.48}& \textbf{57.45}& \textbf{72.38}& \textbf{53.91}&   37.23& \underline{43.15}& \textbf{22.99}& 24.43& \textbf{10.47}& \textbf{9.27}& \textbf{29.65} \\
\bottomrule

\end{tabular}}
\vspace{-0.2in}
\end{table*}

\subsection{Datasets and Metrics}
\paragraph{Datasets}
We conduct experiences on two widely adopted abstractive summarization datasets, including CNN/DailyMail (CNN/DM)~\cite{hermann2015teaching} and Extreme Summarization (XSum)~\cite{narayan2018dont-give}. Both datasets contain news articles and their corresponding summaries written by professional journalists. The number of samples in the train/validation/test set of CNN/DM and XSum is 287,227/13,368/11,490 and 204,045/11,332/11,334, respectively.

\paragraph{Metrics}
Since the traditional evaluation metrics in abstractive summarization, e.g., ROUGE-L (R-L)~\cite{lin2004rouge}, are not capable of measuring the factual consistency between the source document and summary, we adopt several additional metrics for evaluation as follows: (i) \textbf{QAFactEval (QAFE)}~\cite{fabbri2022qafacteval}, which combines entailment and question answering based metrics to capture their complementary signals and further boost the performance. (ii) \textbf{QAGS}~\cite{wang2020asking}, which first generates several questions based on the generated summary with a Question Generation (QG) model, and then generates two sets of corresponding answers given the source document and the summary with a Question Answering (QA) model. Finally, the QAGS score is computed by comparing these answers with token-level similarity metrics. (iii) \textbf{FactCC (FCC)}~\cite{kryscinski2020evaluating}, which is based on a weakly-supervised BERT-based model to measure whether the summary is entailed by the source document. (iv) \textbf{Fact Triple (FT-C/O)}~\cite{goodrich2019assessing}, which extract fact triples (\textit{subject}, \textit{relation}, \textit{object}) separately from the source document and the summary and compare these two sets of triples. Among them, FT-C is in a closed scheme, where \textit{relation} is predicted from a pre-defined relation set; FT-O is in an open scheme, where \textit{relation} is the original text span between \textit{subject} and \textit{object}. (v) \textbf{AVG}, which first calculates the average score over all the factual metrics, and then averages it with the traditional metric R-L for a clear comparison of the trade-off between the traditional and factual metrics.

\subsection{Baselines}
We adopt \textbf{\textsc{Pegasus}}~\cite{zhang2020pegasus} as the model backbone, and mainly choose the following four counterparts to compare with: (i) \textbf{\textsc{Unl}}~\cite{li2020dont}, which leverages the unlikelihood loss to penalize the probabilities of the tokens in unfaithful samples. (ii) \textbf{\textsc{Corr}}~\cite{cao2020factual}, which pre-trains a post-editing corrector model to directly generate factually consistent summaries. (iii) \textbf{\textsc{CCGS}}~\cite{chen2021improving}, which pre-trains a factual consistency predictor and leverages it to rank candidate summaries. (iv) \textbf{\textsc{CLIFF}}~\cite{cao2021cliff}, which adopts contrastive loss to discriminate between faithful and unfaithful samples. (v) \textbf{\textsc{SC}}~\cite{xiao2022entity-based}, which contains an entity-based SpanCopy mechanism with Global Relevance to reduce mismatched entities.

\begin{table}[t]
\caption{Pairwise human evaluation results (\%) in terms of factual consistency compared with \textsc{Pegasus}.}\label{tab:human_eval_fact}
\centering
\setlength\tabcolsep{4pt}
\vspace{-0.1in}
\resizebox{\columnwidth}{!}{
\begin{tabular}{lcccccc}

\toprule
\multirow{2}{*}{\textbf{Methods}}& \multicolumn{3}{c}{\textbf{CNN/DM}}& \multicolumn{3}{c}{\textbf{XSum}} \\
\cmidrule(lr){2-4} \cmidrule{5-7}
~& \textbf{Win$\uparrow$}& \textbf{Tie}& \textbf{Lose$\downarrow$}& \textbf{Win$\uparrow$}& \textbf{Tie}& \textbf{Lose$\downarrow$} \\

\midrule
\textsc{Unl}& 15.33& 54.67& 30.00& \underline{18.00}& 51.33& 30.67 \\
\textsc{Corr}& 13.33& 38.00& 48.67& 7.33& 89.33& \textbf{3.34} \\
\textsc{CCGS}& 8.00& 87.33& \underline{4.67}& 12.67& 78.00& 9.33 \\
\textsc{CLIFF}& \textbf{21.33}& 59.33& 19.34& 17.33& 62.67& 20.00 \\
\textsc{SC}& 14.00& 60.67& 25.33& 6.00& 68.33& 25.67 \\
\midrule
\textsc{CoFactSum}& \underline{17.33}& 80.67& \textbf{2.00}& \textbf{29.33}& 62.00& \underline{8.67} \\
\bottomrule

\end{tabular}}
\vspace{-0.1in}
\end{table}

\section{Implementation Details}
\label{appendix:implementation}
The proposed \textsc{CoFactSum} is implemented based on Huggingface~\cite{wolf2020transformers}.
During the training process in ICT, we set $\gamma, \lambda$ in Equation (\ref{eq:total_loss_cross_attn}) to 1 and 0.01, respectively. The batch size is set to 8, and the number of training steps is set to 50,000 on both datasets. The attending proportion of important/irrelevant information is set to 0.5/0.5 and 0.1/0.9 on CNN/DM and XSum, respectively. The learning rate is set to 5e-4 and 5e-5 on CNN/DM and XSum, respectively. During the training in DDA, the batch size is set to 8, the number of training steps is set to 50,000, and the learning rate is set to 1e-4 on both datasets. And during inference, the masking ratio in ECM is the same as the attending proportion of important information in ICT on both datasets. We use beam search for decoding and set the beam size as 20 and 12 on CNN/DM and XSum, respectively. For the debiasing ratio $\alpha, \beta$ in Equation (\ref{eq:final_eq}), we set $\alpha=0.05, \beta=0.01$ on CNN/DM and $\alpha=0.15, \beta=0.15$ on XSum. The unfaithful samples in DDA are constructed with the system generation method~\cite{cao2021cliff}. All experiments are conducted on GeForce RTX 3090 GPUs.

\begin{table}[t]
\caption{Pairwise human evaluation results (\%) in terms of informativeness compared with \textsc{Pegasus}.}\label{tab:human_eval_inform}
\centering
\setlength\tabcolsep{4pt}
\vspace{-0.1in}
\resizebox{\columnwidth}{!}{
\begin{tabular}{lcccccc}

\toprule
\multirow{2}{*}{\textbf{Methods}}& \multicolumn{3}{c}{\textbf{CNN/DM}}& \multicolumn{3}{c}{\textbf{XSum}} \\
\cmidrule(lr){2-4} \cmidrule{5-7}
~& \textbf{Win$\uparrow$}& \textbf{Tie}& \textbf{Lose$\downarrow$}& \textbf{Win$\uparrow$}& \textbf{Tie}& \textbf{Lose$\downarrow$} \\

\midrule
\textsc{Unl}& 10.33& 59.33& 30.34& \underline{18.67}& 53.33& 28.00 \\
\textsc{Corr}& \underline{12.67}& 67.00& 20.33& 2.33& 95.67& \textbf{2.00} \\
\textsc{CCGS}& 4.00& 93.67& \textbf{2.33}& 2.67& 88.33& \underline{9.00} \\
\textsc{CLIFF}& 10.67& 65.00& 24.33& 10.00& 65.33& 24.67 \\
\textsc{SC}& \textbf{14.33}& 66.67& 19.00& 12.33& 66.33& 21.34 \\
\midrule
\textsc{CoFactSum}& 8.33& 83.00& \underline{8.67}& \textbf{22.00}& 65.67& 12.33 \\
\bottomrule

\end{tabular}}
\vspace{-0.1in}
\end{table}

\subsection{Results}
\paragraph{Automatic Evaluation}
We report the automatic evaluation results on CNN/DM and XSum in Table~\ref{tab:main_results}. Following previous studies~\cite{cao2021cliff}, we randomly select 5,000 samples for the factual consistency evaluation on CNN/DM.
In summary, the overall performances of \textsc{CoFactSum} on both CNN/DM and XSum are significantly better than baseline models with improvements of at least 1.10\% and 0.31\%, respectively. 
Specifically, in terms of most of the factual consistency metrics, the proposed \textsc{CoFactSum} shows advantages compared with baselines. For example, evaluated by QAFE, QAGS, and FCC on CNN/DM, \textsc{CoFactSum} achieves 0.93\%, 0.42\%, and 4.05\% improvements compared with \textsc{Pegasus}, respectively; and evaluated by QAFE, FT-C, and FT-O on XSum, \textsc{CoFactSum} achieves 1.66\%, 4.30\% and 5.55\% improvements over \textsc{Pegasus}, respectively.
On the other hand, although a slight drop happens on the traditional metric R-L (similar to previous studies CCGS and CLIFF), the proposed \textsc{CoFactSum} achieves competitive performances compared with baselines, which confirms the informativeness of summaries generated by the proposed method.

\paragraph{Human Evaluation}
We further perform pairwise human evaluations on the informativeness and factual consistency, as shown in Table~\ref{tab:human_eval_fact} and \ref{tab:human_eval_inform}. 
We randomly select 100 samples in CNN/DM and XSum, and invite three experienced annotators to judge whether the summaries generated by the factually consistent methods are {\it better than}, {\it tie with}, or {\it worse than} those generated by baseline \textsc{Pegasus}. 
From the results in Table~\ref{tab:human_eval_fact}, we can see that compared with baseline methods, \textsc{CoFactSum} has the fewest lose samples with a percentage of 2.00\% in CNN/DM and the most win samples with a percentage of 29.33\% in XSum, demonstrating that the proposed method significantly improves the factual consistency in generated summaries.
The results in Table~\ref{tab:human_eval_inform} show that \textsc{CoFactSum} are competitive with baseline methods, indicating the proposed \textsc{CoFactSum} achieves a great balance between informativeness and factual consistency.

\begin{table}[t]
\caption{Ablation study on different modules.}\label{tab:ablation_module}
\centering
\vspace{-0.1in}
\resizebox{0.95\columnwidth}{!}{
\begin{tabular}{lcgggc}

\toprule
\textbf{Methods}& \textbf{R-L}& \textbf{QAGS}& \textbf{FT-C}& \textbf{FT-O}& \textbf{AVG} \\
\midrule
Ours& 37.23& \textbf{23.44}& \textbf{9.84}& \textbf{8.96}& \textbf{25.66} \\
\midrule
\textit{w/o} DDA&  37.64& 22.79& 8.00& 7.68& 25.23 \\
\textit{w/o} ECM&  37.89& 22.95& 7.29& 7.70& 25.27 \\
\textit{w/o} ICT&  38.50& 21.68& 5.97& 4.36& 24.59 \\
\textit{w/o} All&  \textbf{39.06}& 21.29& 5.71& 3.77& 24.66 \\
\bottomrule

\end{tabular}}
\vspace{-0.1in}
\end{table}

\subsection{Ablation Study}
We conduct an ablation study to evaluate the effectiveness of different modules (including DDA, ECM, and ICT) of \textsc{CoFactSum}. We randomly select 3,000 instances from XSum for evaluation, and the experimental results are shown in Table~\ref{tab:ablation_module}.

From the table, we can observe that the overall performances all decrease when removing any one module of \textsc{CoFactSum}. 
Specifically, the DDA and ECM modules contribute almost equally to the overall performance with average improvements of 0.43\% and 0.39\%, respectively, and the ICT module contributes the most to the overall performance with an improvement of 1.07\%. 
These experimental results demonstrate the effectiveness of the modules of \textsc{CoFactSum}. 

Besides, we conduct an ablation study to evaluate the effectiveness of the adopted training losses (including $\mathcal{L}_{unl}$, $\mathcal{L}_{xent}$, and $\mathcal{L}_{kl}$) in \textsc{CoFactSum} and show the experimental results in Table~\ref{tab:ablation_loss}.  
From the table, we can observe that the improvements brought by the adopted training losses in \textsc{CoFactSum} are also significant. For example, the KL loss $\mathcal{L}_{kl}$ achieves the greatest promotion to the overall performance with an improvement of 0.82\%. 

\begin{table}[t]
\caption{Ablation study on different training losses.}\label{tab:ablation_loss}
\centering
\vspace{-0.1in}
\resizebox{0.95\columnwidth}{!}{
\begin{tabular}{lcgggc}

\toprule
\textbf{Methods}& \textbf{R-L}& \textbf{QAGS}& \textbf{FT-C}& \textbf{FT-O}& \textbf{AVG} \\
\midrule
Ours& 37.23& \textbf{23.44}& \textbf{9.84}& \textbf{8.96}& \textbf{25.66} \\
\midrule
\textit{w/o} $\mathcal{L}_{unl}$&   \textbf{38.46}& 22.24& 6.98& 6.87& 25.25 \\
\textit{w/o} $\mathcal{L}_{xent}$& 37.85& 22.45& 7.17& 6.27& 24.91 \\
\textit{w/o} $\mathcal{L}_{kl}$&   38.08& 21.93& 7.20& 5.67& 24.84 \\
\bottomrule

\end{tabular}}
\vspace{-0.1in}
\end{table}

\subsection{Further Discussions}
In this section, we provide more analysis and discussions to gain a deeper understanding of the debiasing mechanisms used in \textsc{CoFactSum}.

\begin{figure}
\includegraphics[trim=10 0 10 0, width=0.92\columnwidth, clip]{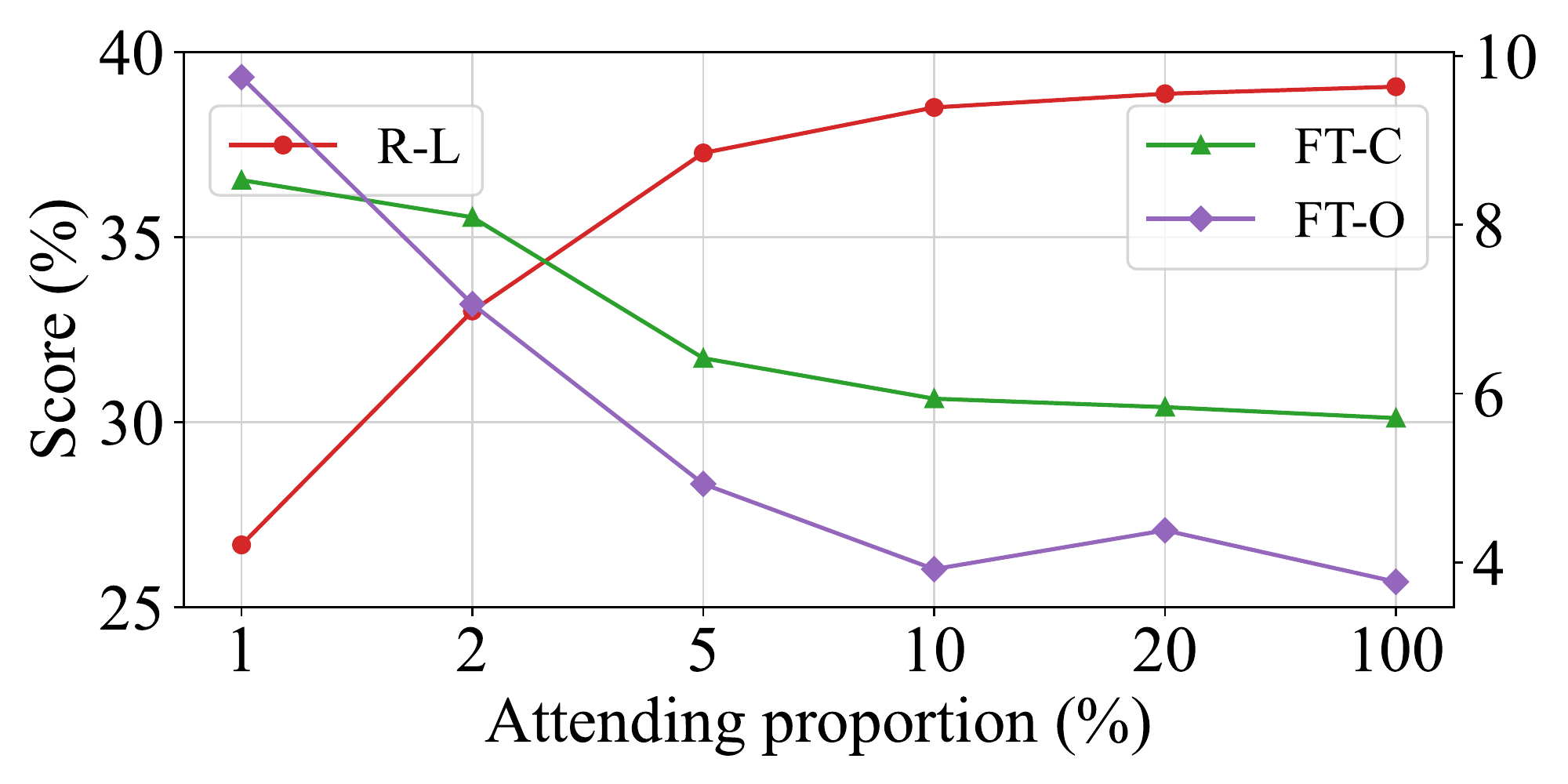}
\vspace{-0.15in}
\centering
\caption{Analysis on different attending proportions of irrelevant information.}\label{fig:intro}
\vspace{-0.05in}
\end{figure}

\paragraph{Impact of Irrelevancy Bias}
We conduct several experiments on the original \textsc{Pegasus} model to evaluate the negative impact of irrelevancy bias on factual consistency. Specifically, we force the model attends to different proportions of irrelevant information based on the cross-attention scores during decoding and assess the generated summaries. The results are shown in Figure~\ref{fig:intro}, from which we can observe that the factual consistency scores (i.e., FT-C and FT-O) gradually decrease as the attending proportion and the amount of irrelevant information increase, demonstrating the negative effect of irrelevancy bias.

\paragraph{Masking and Attending Strategy in ECM and ICT}
To confirm the ascendancy of the dynamic strategy, we select several static strategies for comparison. 
Following previous works that adopt static masking strategies~\cite{xie2021factual}, we select three static types for masking and attending in ECM and ICT, including \textit{token-level} (tok.), \textit{sentence-level} (sent.), and \textit{document-level} (doc.). 
These strategies are proposed to mask and attend to the same named entities, the same sentences with at least one entity, and the entire tokens in the source document during different decoding steps in ECM and ICT, respectively. 
The results are shown in Table~\ref{tab:analysis_mask}, from which we observe that all the overall performances of the static strategies have significant decreases compared with the proposed dynamic strategy used in \textsc{CoFactSum}. 
Moreover, we can see that the strategies \textit{token-level} and \textit{sentence-level} lead to poor performances on R-L while those of \textit{document-level} and the dynamic strategy are kept at the same level.
These results imply that the decoder has different perceptions of important information at different decoding steps; simply choosing the same part of the source document as important information will harm the informativeness, while indiscriminately choosing the entire tokens or dynamically choosing the important tokens can alleviate the issue.

\begin{table}[h]
\caption{Analysis on different masking and attending strategies in ECM and ICT.}\label{tab:analysis_mask}
\vspace{-0.1in}
\centering
\resizebox{0.95\columnwidth}{!}{
\begin{tabular}{lcgggc}

\toprule
\textbf{Methods}& \textbf{R-L}& \textbf{QAGS}& \textbf{FT-C}& \textbf{FT-O}& \textbf{AVG} \\
\midrule
Ours& 37.23& \textbf{23.44}& \textbf{9.84}& \textbf{8.96}& \textbf{25.66} \\
\midrule
Static (tok.)&  32.93& 21.51& 9.66& 8.54& 23.08 \\
Static (sent.)& 35.34& 20.51& 7.90& 5.30& 23.29 \\
Static (doc.)&  \textbf{38.08}& 20.95& 7.77& 4.80& 24.63 \\
\bottomrule

\end{tabular}}
\vspace{-0.1in}
\end{table}

\begin{figure}
\includegraphics[trim=10 0 10 0, width=0.92\columnwidth, clip]{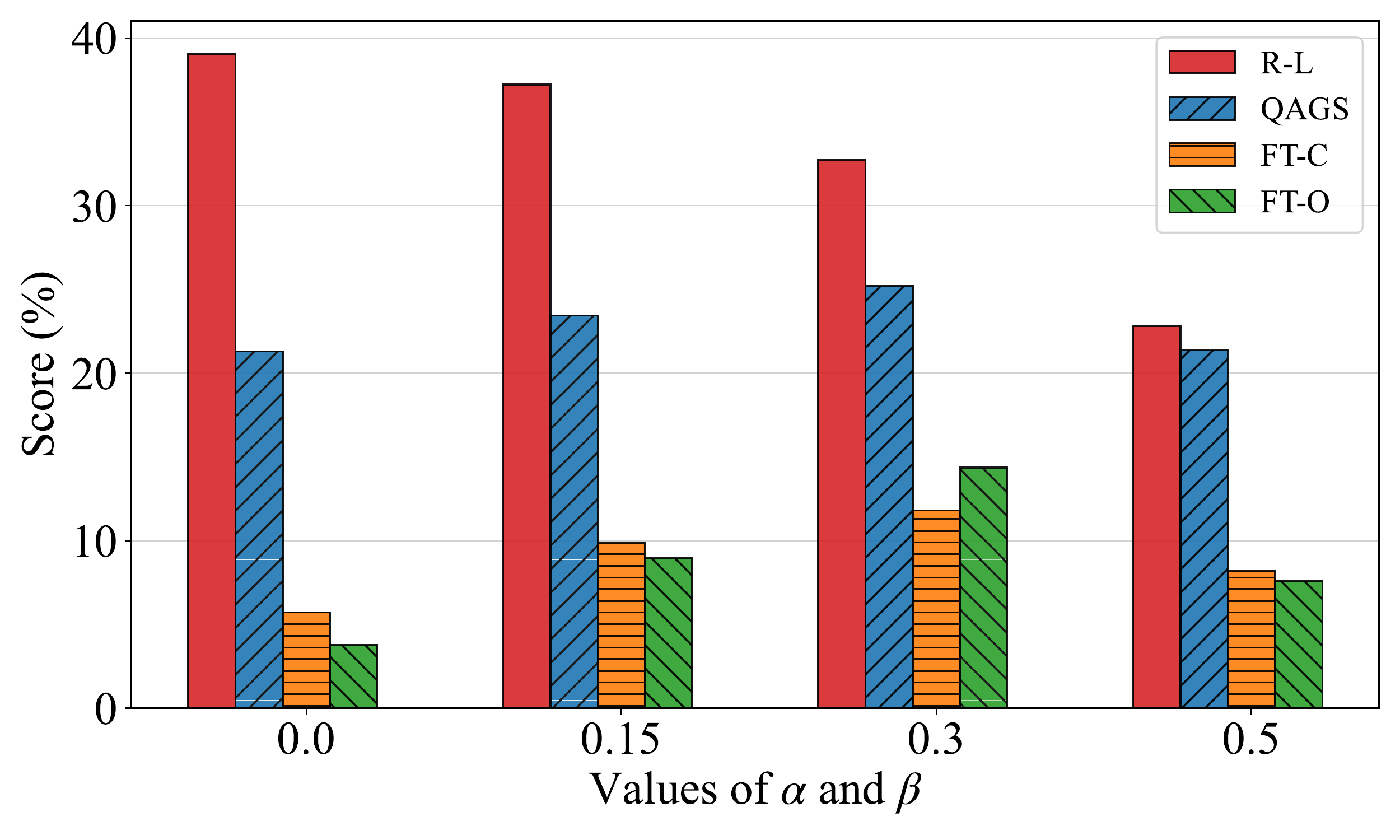}
\vspace{-0.15in}
\centering
\caption{Comparisons among applying different debiasing ratios $\alpha$ and $\beta$ in \textsc{CoFactSum}.}\label{fig:different_a_b}
\vspace{-0.05in}
\end{figure}

\paragraph{Impact of Debiasing Degree}
We gradually increase the static debiasing ratio $\alpha, \beta$ in Equation (\ref{eq:final_eq}) to investigate the impact of debiasing degree on the informativeness and factual consistency. From the results in Figure~\ref{fig:different_a_b}, we can see that with the enhancement of debiasing degree, the R-L score gradually decreases and the factual consistency scores increase first and then decrease. This phenomenon indicates that a proper debiasing degree can improve the factual consistency of generated summaries without weakening their informativeness, while a large debiasing degree might severely hurt the informativeness and factual consistency. 

\paragraph{Case Study}
We further conduct a case study in Figure~\ref{fig:case_study}. 
From the figure, we can conclude that all the summaries generated by the baselines contain inconsistent facts that are not mentioned in the source document, such as the number of employees ``100 workers'' and the name of the city ``Cardiff'', while the proposed \textsc{CoFactSum} alleviates such factual inconsistency issue to some extent. 

\section{Related Work}
\paragraph{Factual Consistency in Text Summarization}
As discussed in previous studies~\cite{maynez2020on-faithfulness,goyal2021annotating,nan2021entity-level,ladhak2022faithful,ribeiro2022factgraph}, current advanced generation models in abstractive summarization are prone to produce factually inconsistent texts that contain contradictory facts against the source documents. 
To tackle such issues, many techniques have been proposed recently, which can be mainly divided into three categories. The first is \textit{fact encoding}, which aims to incorporate more fact-related information during encoding source documents or decoding target summaries, such as fact descriptions~\cite{cao2018faithful}, entailment knowledge~\cite{li2018ensure}, knowledge graphs~\cite{huang2020knowledge,zhu2021enhancing}, and document entities~\cite{xiao2022entity-based}. The second is \textit{post editing}, which treats the generated summaries as drafts and further conducts post-editing on them, and is usually achieved by a separate correction model~\cite{dong2020multi-fact,cao2020factual,chen2021improving}. The third is \textit{auxiliary loss applying}, which designs auxiliary penalty losses to force the model to distinguish between faithful and unfaithful samples, and so far, the unlikelihood loss~\cite{li2020dont} and contrastive loss~\cite{cao2021cliff,liu2022co2sum,wan2022factpegasus} are most widely adopted.
Differently, we aim to enhance the factual consistency with counterfactual estimation, which is able to discover the intrinsic causes and remove their negative effects based on their causal relationships.

\begin{table}
\centering
\resizebox{\columnwidth}{!}{
\begin{tabularx}{1.25\columnwidth}{p{1.2\columnwidth}}
\toprule
\textbf{Source document}: \textcolor{myblue}{The employees, who worked in four takeaways, are alleged to have been living and working in the country illegally.} The firms have been asked to produce documents proving their staff had the right to work and live in the \textcolor{myblue}{UK}. If they are unable to do so \textcolor{myblue}{the Home Office} said they would impose a fine of up to £20,000 per illegal employee. \textcolor{myblue}{The process to deport the workers is already under way.} \\

\midrule
\textbf{\textsc{Pegasus}}~\cite{zhang2020pegasus}: The Home Office has launched an investigation into the alleged illegal employment of \textcolor{myred}{more than 100 workers} at takeaways \textcolor{myred}{in Cardiff}. \\
\textbf{\textsc{CCGS}}~\cite{chen2021improving}: The Home Office has launched an investigation into the alleged illegal employment of \textcolor{myred}{four workers} at takeaways \textcolor{myred}{in Cardiff}. \\
\textbf{\textsc{CLIFF}}~\cite{cao2021cliff}: \textcolor{myred}{More than 100 illegal workers} have been ordered to leave the UK by the Home Office. \\
\textbf{\textsc{CoFactSum} (Ours)}: The Home Office has launched an operation targeting illegal immigrants working in the takeaway food industry. \\

\bottomrule
\end{tabularx}}
\vspace{-0.05in}
\captionof{figure}{An example of generated summaries by baselines and \textsc{CoFactSum}. The \textcolor{myblue}{supporting facts} in the source document and \textcolor{myred}{inconsistent facts} in the generated summaries are marked in \textcolor{myblue}{blue} and \textcolor{myred}{red}, respectively.}\label{fig:case_study}
\vspace{-0.1in}
\end{table}

\paragraph{Counterfactual Inference}
In the field of natural language processing, causal inference~\cite{pearl2001direct,pearl2018the-book} has recently inspired many works to discover the intrinsic causes of specific biases and remove their causal effects in an interpretable way, such as the studies in visual question answering~\cite{niu2021counterfactual-vqa}, text classification~\cite{qian2021counterfactual}, gender bias~\cite{vig2020investigating}, and text summarization~\cite{xie2021factual}. These methods target measuring the causal effects of biases under counterfactual scenarios based on pre-defined causal graphs and eliminating their causal effects by mitigating them from the total effect. In this study, we incorporate the method into the generation process of summarization to improve factual consistency of generated summaries.

\section{Conclusion}
In this paper, to enhance the factual consistency of generated summaries, we aim to alleviate the language bias and irrelevancy bias in abstractive text summarization via counterfactual estimation. 
Specifically, we propose \textsc{CoFactSum}, a debiasing framework to remove the causal effects of the biases with two types of counterfactual estimation methods, including Explicit Counterfactual Masking and Implicit Counterfactual Training. 
Meanwhile, we design a Debiasing Degree Adjustment module to dynamically adjust the debiasing degrees at different decoding steps. 
Experimental results on two widely-used summarization datasets demonstrate the superiority of the proposed \textsc{CoFactSum} in improving factual consistency.

\bibliography{acl}
\bibliographystyle{acl_natbib}

\end{document}